\documentclass{applemlr}
\usepackage{amsmath}
\usepackage{enumerate}
\usepackage{algorithm}
\usepackage{algpseudocode}
\usepackage{amsfonts}
\usepackage{amsthm}
\usepackage{cleveref}
\usepackage{diagbox}
\usepackage{colortbl}
\usepackage{amssymb}
\usepackage{xspace}
\usepackage{wrapfig}
\usepackage{adjustbox}
\usepackage{tabularx}
\usepackage{booktabs}
\usepackage{mathtools}
\usepackage{tikz}
\usepackage{enumitem}
\usepackage{silence}
\usepackage{dsfont}
\usepackage[table]{xcolor}
\usepackage[dvipsnames]{xcolor}
\usepackage{multirow}
\usepackage{makecell}
\usepackage{xfakebold}
\usepackage{twemojis}

\usepackage[table]{xcolor}

\definecolor{dgreen}{rgb}{0.0, 0.5, 0.0}                                                               \definecolor{dred}{rgb}{0.8, 0.0, 0.0}
\usepackage{xcolor, multirow, booktabs}

\usepackage{amsmath,amsfonts,bm}

\def\eqref#1{equation~\ref{#1}}

\def\1{\bm{1}}

\DeclareMathAlphabet{\mathsfit}{\encodingdefault}{\sfdefault}{m}{sl}
\SetMathAlphabet{\mathsfit}{bold}{\encodingdefault}{\sfdefault}{bx}{n}

\definecolor{textgray}{HTML}{6E6E73}
\usetikzlibrary{positioning, calc}
\usetikzlibrary{decorations.pathmorphing}

\makeatletter
\patchcmd{\wrong@fontshape}{\@gobbletwo}{}{}{}
\makeatother
\numberwithin{equation}{section}
\makeatletter
\AtBeginDocument{
  \urlstyle{sf}
  
}
\makeatother

\definecolor{light}{RGB}{125, 125, 125}
\crefname{tcb@cnt@pbox}{code}{code}
\Crefname{tcb@cnt@pbox}{Code}{Code}
\crefname{assumption}{assumption}{assumption}
\Crefname{assumption}{Assumption}{Assumptions}

\newtcolorbox[auto counter]{pbox}[2][]{
  colback=white,
  title=Code~\thetcbcounter: #2,
  #1,fonttitle=\sffamily,
  fontupper=\sffamily,
  arc=2pt,
  colframe=bgcolor,
  coltitle=fgcolor,
  colbacktitle=bgcolor,
  toptitle=0.25cm,
  bottomtitle=0.125cm
}

\makeatletter
\newcommand\applefootnote[1]{%
  \begingroup
  \renewcommand\thefootnote{}%
  \renewcommand\@makefntext[1]{\noindent##1}%
  \footnote{#1}%
  \addtocounter{footnote}{-1}%
  \endgroup
}
\makeatother

\definecolor{cverbbg}{gray}{0.90}

\title{BalCapRL\scalebox{2.5}{\twemoji{billed cap}}: A Balanced Framework for RL-Based MLLM Image Captioning}

\author[1]{Shaokai Ye}
\author[1]{Vasileios Saveris}
\author[1]{Yihao Qian}
\author[1]{Jiaming Hu}
\author[1]{Elmira Amirloo}
\author[1]{Peter Grasch}

\affiliation[1]{Apple}

\correspondence{
  \email{shaokai\_ye@apple.com},
  \email{pgrasch@apple.com}
}
\date{\sffamily\today}

\abstract{Image captioning is one of the most fundamental tasks in computer vision. Owing to its open-ended nature, it has received significant attention in the era of multimodal large language models (MLLMs). In pursuit of ever more detailed and accurate captions, recent work has increasingly turned to reinforcement learning (RL). However, existing captioning-RL methods and evaluation metrics often emphasize a narrow notion of caption quality, inducing trade-offs across core dimensions of captioning. For example, utility-oriented objectives can encourage noisy, hallucinated, or overlong captions that improve downstream question answering while harming fluency, whereas arena-style objectives can favor fluent but generic descriptions with limited usefulness. To address this, we propose a more balanced RL framework that jointly optimizes utility-aware correctness, reference coverage, and linguistic quality. In order to effectively optimize the resulting continuous multi-objective reward formulation, we apply GDPO-style reward-decoupled normalization to continuous-valued captioning rewards and show that it improves performance over vanilla GRPO. Additionally, we introduce length-conditional reward masking, yielding a more suitable length penalty for captioning. Across LLaVA-1.5-7B and Qwen2.5-VL 3B and 7B base models, our method consistently improves caption quality, with peak gains of +13.6 DCScore, +9.0 CaptionQA, and +29.0 CapArena across different models.}

\begin{document}
\maketitle

\raggedbottom

\section{Introduction}\label{sec:intro}

Image captioning is a fundamental visual task. Early captioning models tend to generate short descriptions centered around closed-vocabulary objects. Advances in MLLMs~\citep{liu2024improvedbaselinesvisualinstruction, qwen2025qwen25technicalreport} have enabled increasing open-ended and detailed captions.

\begin{figure}[t]
    \centering
    \includegraphics[width=0.9\linewidth]{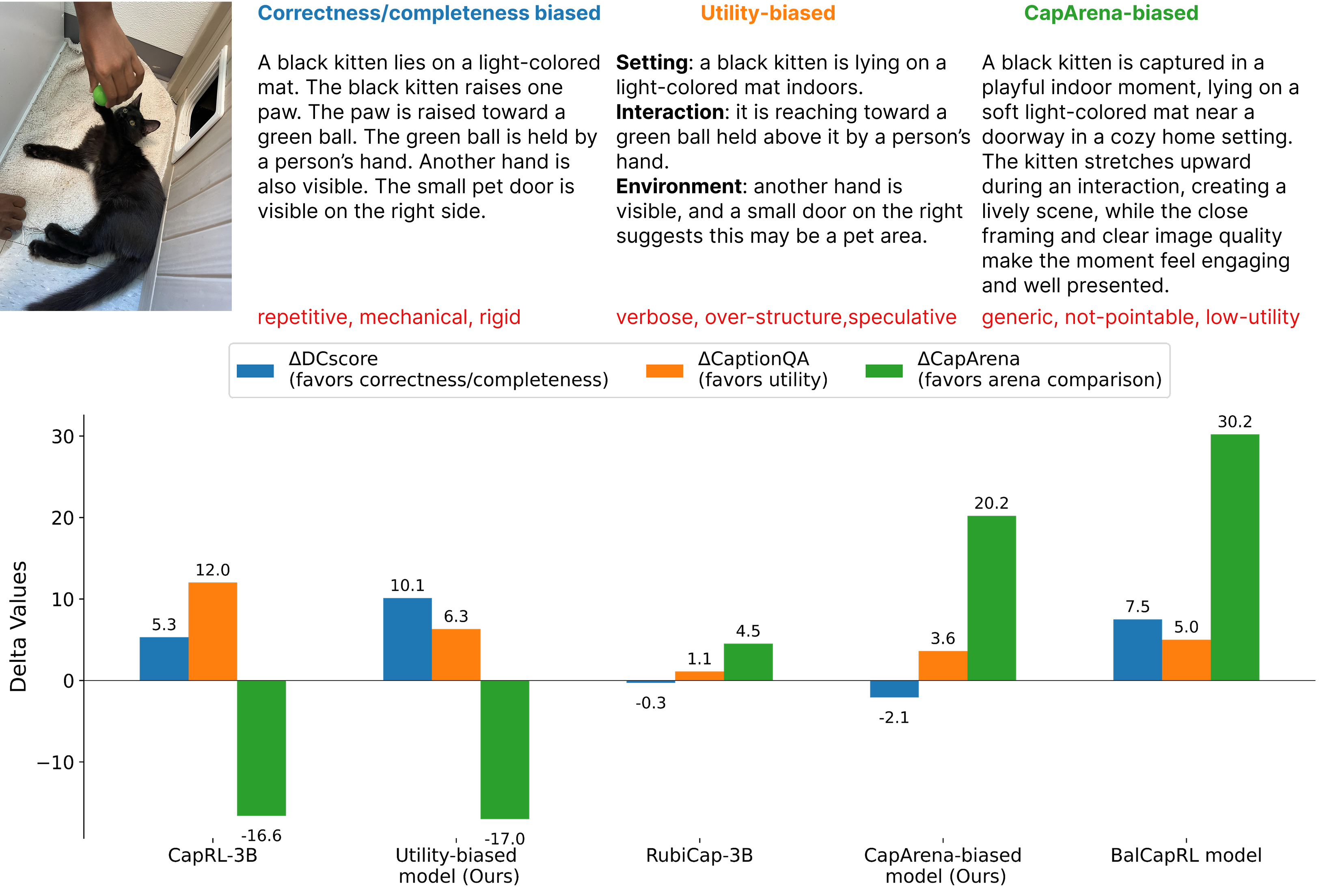}
\caption{\textbf{Top.} Illustration of captions from different biased models. \textbf{Bottom.} Different captioning-RL models evaluated on benchmarks representing different views. $\Delta$ (higher is better) denotes the difference between the RL-trained model and the base QwenVL2.5-3B model. CapRL and RubiCap results are produced by evaluating official checkpoints.}
    \label{fig:Fig1}
\end{figure}

In order to maximize captioning capabilities of modern MLLMs, reinforcement learning aimed at improving captioning performance as an objective (captioning-RL)~\citep{xing2025caprlstimulatingdenseimage, huang2026rubicaprubricguidedreinforcementlearning, ye2025paintingwordselevatingdetailed} has increasingly gained popularity. 

Existing captioning-RL methods often optimize a narrow notion of caption quality and then evaluate improvements using benchmarks aligned with that same perspective. We find that this creates a systematic bias: gains on one dimension of caption quality often come with regressions or moderate improvements on others.

We identify three major views that currently shape captioning-RL and caption evaluation: downstream utility~\citep{yang2025captionqacaptionusefulimage}, correctness-and-completeness with respect to reference captions~\citep{ye2025paintingwordselevatingdetailed}, and arena-style preference judgments~\citep{cheng2025caparenabenchmarkinganalyzingdetailed}. Each captures an important aspect of caption quality, but optimizing any one view in isolation is insufficient. Correctness-and-coverage objectives can reward repetitive, mechanical and rigid descriptions. Utility-oriented training can encourage hallucinated or overly long captions that help downstream question answering while degrading fluency. Arena-style judgments, in contrast, can favor fluent yet generic captions that rank well in CapArena despite being less useful and less informative. As a result, prior methods often exhibit clear trade-offs across benchmarks rather than uniformly better captioning. Figure~\ref{fig:Fig1} illustrates this pattern for representative prior methods, as well as purposefully biased variants of our method, obtained by removing individual components from our framework.
To address this issue, we propose BalCapRL, a more balanced reinforcement learning framework for detailed image captioning. Our method jointly optimizes rewards for utility-aware correctness, reference-coverage completeness, and linguistic quality. Because these reward dimensions can have distinct and partially competing optimization dynamics, we find that vanilla GRPO is suboptimal in our setting, and therefore apply GDPO~\citep{liu2026gdpogrouprewarddecouplednormalization} to continuous-valued rewards, which we refer to as c-GDPO, to better optimize multi-reward policy optimization (Figure~\ref{fig:gdpo_illustration}). Additionally, we introduced a novel two-sided length penalty via reward masking, which we show is better suited to captioning-RL.

Across LLaVA-1.5-7B~\citep{liu2024improvedbaselinesvisualinstruction} and QwenVL2.5 3B and 7B~\citep{qwen2025qwen25technicalreport}, BalCapRL consistently improves caption quality across benchmarks representing all three views, outperforming prior methods in almost all settings. These results suggest that better captioning-RL requires not just optimization toward a single benchmark, but a more balanced training objective that explicitly accounts for multiple dimensions of caption quality.
\section{Method}\label{sec:method}

\subsection{Reward design}

To obtain scalar reward signals for caption quality, we first incorporate the correctness-and-completeness perspective. Our method is related in spirit to FEEDQUILL~\citep{ye2025paintingwordselevatingdetailed} in that both decompose captions into atomic assertions for rewards derived from precision and recall. Specifically, we decompose both the policy generated caption as well as a ground-truth caption into atomic assertions to enable the computation of precision and recall, which provide reward signals for correctness and completeness, respectively. Unlike FEEDQUILL, our method does not require training separate reward models; instead, we compute the rewards directly via judge-based decomposition and verification, yielding a simpler and more modular pipeline. However, as discussed in Section~\ref{sec:intro}, correctness and completeness alone are insufficient: they do not prevent captions from being correct yet not useful, nor do they prevent degradation in fluency. We therefore introduce two additional components: a pointability principle, used as a rubric to constrain what is considered as a useful atomic assertion, and a linguistic score that regularizes the model toward fluent and coherent captions.

\textbf{Decomposition.} Given a model-generated caption $C$, we employ a large language model (LLM) to decompose it into a set of \textbf{atomic assertions}:
\begin{equation}
\mathcal{A} = \{a_1, a_2, \ldots, a_N\},
\end{equation}
where $N$ denotes the total number of atomic assertions extracted from the caption. Similarly, the reference caption (data generation details in Appendix~\ref{sec:implementation_settings}) is decomposed into a set of reference units:
\begin{equation}
\mathcal{O} = \{o_1, o_2, \ldots, o_M\},
\end{equation}
where $M$ is the number of reference units.

\textbf{Precision (Utility-aware Correctness).} The precision reward $R_{\mathrm{prec}}$ measures the proportion of atomic assertions in $\mathcal{A}$ that are \textbf{verifiably correct}. An atomic assertion $a_i \in \mathcal{A}$ is considered a \textbf{true positive} if and only if it satisfies the following two conditions:
\begin{enumerate}
    \item \textbf{Visually verifiable} ($\mathcal{A}_G$): The assertion can be  verified as factually correct from the image content by a vision-language model (VLM).
    \item \textbf{Pointability} ($\mathcal{A}_P$): The assertion refers to a visually pointable element---specifically, something that a person can physically point to in the image. Compared with prior work~\citep{ye2025paintingwordselevatingdetailed}, this novel addition specifically discourages generating non-pointable, low utility meta commentary. An empirical example is shown in~Figure~\ref{fig:pointability_qualitative_example} and the prompt is provided in Appendix~\ref{sec:training_prompt}.
\end{enumerate}

Formally, let $\mathcal{A}^+ \subseteq \mathcal{A}$ denote the set of true positive assertions:
\begin{equation}
\mathcal{A}^+ = \mathcal{A}_G \cap \mathcal{A}_P,
\end{equation}
where $\mathcal{A}_G = \{a_i \in \mathcal{A} \mid a_i \text{ is visually verified}\}$ and $\mathcal{A}_P = \{a_i \in \mathcal{A} \mid a_i \text{ is pointable}\}$.

The precision reward is then computed as: $R_{\mathrm{prec}} = \frac{|\mathcal{A}^+|}{|\mathcal{A}|}$.

\textbf{Recall (Reference Coverage).} The recall reward $R_{\mathrm{rec}}$ measures the extent to which the model-generated caption covers the key information present in the reference caption. We employ an LLM to perform the matching, assessing whether each atomic assertion $o_j \in \mathcal{O}$ is mentioned or can be reasonably inferred from the generated atomic assertions (Appendix \ref{sec:training_prompt}). 

Let $\mathcal{Q} = \mathcal{A} \cap \mathcal{O}$ denote the set of matched units between the generated and reference captions, as determined by the LLM. The recall reward is computed as $R_{\mathrm{rec}} = \frac{|\mathcal{Q}|}{|\mathcal{O}|}$.

\textbf{Linguistic Score.} The linguistic reward $R_{\mathrm{ling}}$ evaluates the linguistic quality of generated captions using an LLM (Appendix~\ref{sec:training_prompt}) that assesses three dimensions: \textit{Clarity}, measuring readability and absence of ambiguity; \textit{Fluency}, evaluating grammatical correctness and natural phrasing; and \textit{Coherency}, assessing logical flow and unified structure. Each of the three dimensions is normalized to the range [0,1], and the final linguistic reward is computed as their average.


\subsection{Data}


Across experiments, we use images from ShareGPT4V~\citep{chen2023sharegpt4vimprovinglargemultimodal}. The dataset contains roughly 90K image-text pairs, originally captioned by GPT-4V~\citep{openai2024gpt4technicalreport}. We re-captioned the data with GPT-5-mini~\citep{singh2025openaigpt5card}, reusing the original captioning prompts, and use these updated reference captions for our main results.

\subsection{Policy Optimization}

\textbf{Applying GDPO to Continuous Captioning Rewards.}
Recently, GRPO~\citep{shao2024deepseekmathpushinglimitsmathematical} and its variants~\citep{liu2026gdpogrouprewarddecouplednormalization, yu2025dapoopensourcellmreinforcement, gao2025softadaptivepolicyoptimization,liu2025understandingr1zeroliketrainingcritical} have become widely used policy optimization methods. In the original GRPO and most of its follow-ups, when multiple rewards are present, these rewards are first summed and then group-normalized, which can lead to the collapse of distinct rollout advantages~\citep{liu2026gdpogrouprewarddecouplednormalization}. GDPO~\citep{liu2026gdpogrouprewarddecouplednormalization} addresses this issue by decoupling normalization across reward dimensions. 

We observe that this pathology is not limited to discrete or verifiable reward settings. In our setting, the precision, recall, and linguistic rewards are continuous-valued with distinct dynamics. Nevertheless, vanilla GRPO still sums these rewards before group normalization, reducing each rollout to a single scalar. As a result, the resulting advantage depends only on a one-dimensional projection of the reward vector, causing distinct continuous reward trade-offs to become indistinguishable when their aggregated rewards coincide. Thus:

\begin{figure}[t]
    \centering        \includegraphics[width=0.8\linewidth]{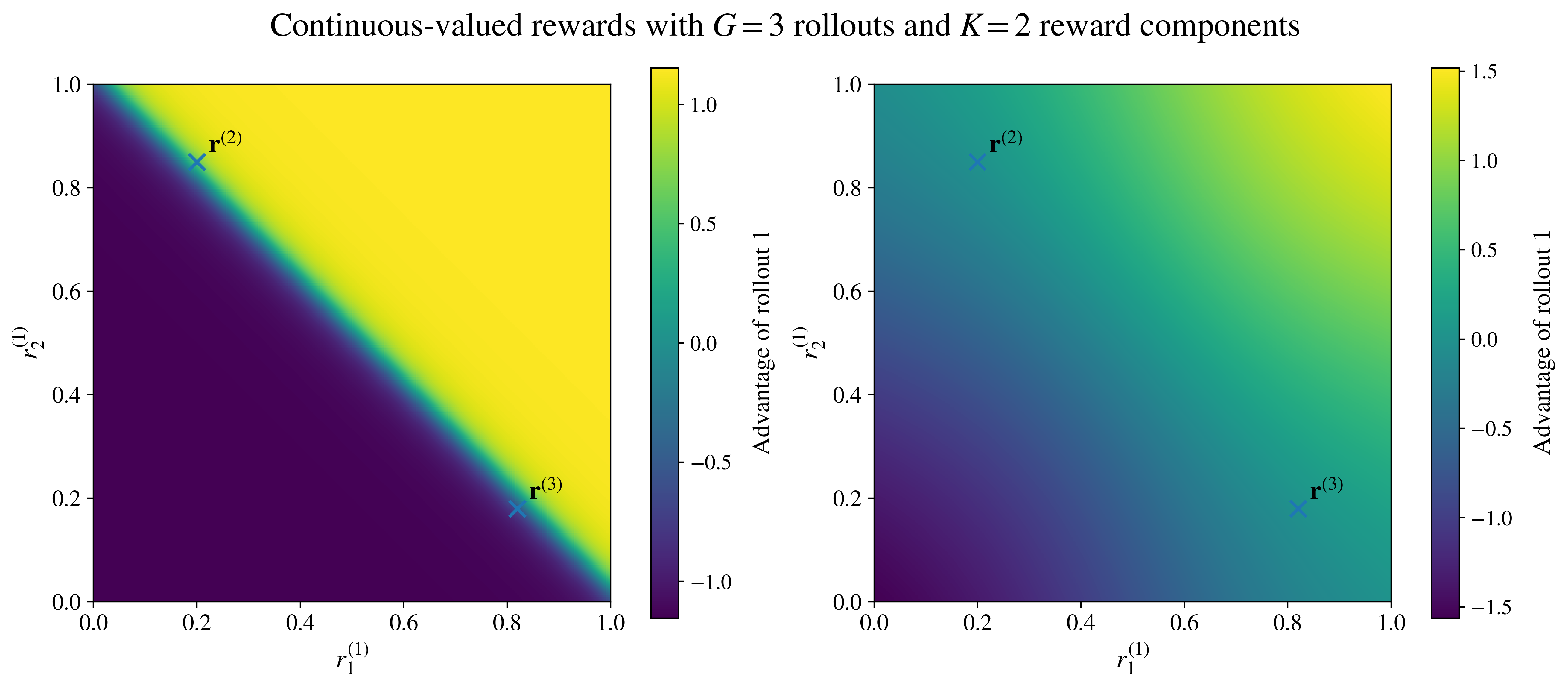}
    \caption{\textbf{Illustration of summed-reward collapse under vanilla GRPO.}
Left: when rewards are aggregated before group normalization, the normalized advantage depends only on the aggregated reward, so reward vectors with identical aggregates are indistinguishable, and nearby aggregates may be only weakly separated. Right: c-GDPO normalizes rewards separately before aggregation, avoiding this invariance and yielding more distinguishable advantages in the same setting. We vary rollout 1 over $[0,1]^2$ while fixing $r^{(2)}=(0.20,0.85)$ and $r^{(3)}=(0.82,0.18)$.}
    \label{fig:gdpo_illustration}
\end{figure}    

\textbf{Proposition 1.} \textit{Consider a $K$-reward, $G$-rollout setting with continuous-valued rewards, where
$G \ge 3$ and each reward dimension has nonzero within-group variance. Under vanilla GRPO, if rewards are aggregated before group normalization, then for any fixed competing rollouts, the normalized advantage of a rollout depends on its reward vector only through its aggregated reward. Consequently, all reward vectors lying on the same aggregated-reward hyperplane are indistinguishable to the optimizer. In contrast, reward-decoupled normalization computes per-reward normalized deviations before aggregation, and therefore is not invariant to these hyperplanes.}

\textit{Proof} is provided in Appendix~\ref{sec:full_proof}.

Therefore, following the same decoupled-normalization principle as GDPO, we apply it to continuous-valued multi-reward optimization, which we refer to this continuous-reward instantiation as c-GDPO. This enables its application to our precision, recall, and linguistic rewards, while preserving finer distinctions among different reward combinations and providing more expressive training signals.

To illustrate the effect of c-GDPO in the continuous-valued setting, Figure~\ref{fig:gdpo_illustration} shows that vanilla GRPO loses fine-grained multi-reward signal after reward aggregation and normalization. In particular, in the saturated region, different underlying reward combinations can yield nearly identical advantage values. In contrast, c-GDPO preserves these differences in the final advantage (Figure~\ref{fig:gdpo_illustration}), which leads to more stable optimization in our setting. 

Specifically, given a batch of $G$ rollouts for each input, we first compute the normalized advantage for each reward. For the $j$-th rollout, the individual advantages are:
\begin{equation}
A^{(j)}_{\mathrm{prec}} = \frac{R^{(j)}_{\mathrm{prec}} - \mu_{\mathrm{prec}}}{\sigma_{\mathrm{prec}}}, \quad
A^{(j)}_{\mathrm{rec}} = \frac{R^{(j)}_{\mathrm{rec}} - \mu_{\mathrm{rec}}}{\sigma_{\mathrm{rec}}}, \quad
A^{(j)}_{\mathrm{ling}} = \frac{R^{(j)}_{\mathrm{ling}} - \mu_{\mathrm{ling}}}{\sigma_{\mathrm{ling}}},
\end{equation}
where $\mu_{\mathrm{prec}}$, $\mu_{\mathrm{rec}}$, $\mu_{\mathrm{ling}}$ and $\sigma_{\mathrm{prec}}$, $\sigma_{\mathrm{rec}}$, $\sigma_{\mathrm{ling}}$ denote the mean and standard deviation of each respective reward across all $G$ rollouts in the group.

The overall advantage is then obtained by a weighted sum of the normalized advantages:
\begin{equation}
A^{(j)}_{\mathrm{sum}} = w_{\mathrm{prec}} A^{(j)}_{\mathrm{prec}} + w_{\mathrm{rec}} A^{(j)}_{\mathrm{rec}} + w_{\mathrm{ling}} A^{(j)}_{\mathrm{ling}},
\end{equation}
where $w_{\mathrm{prec}}$, $w_{\mathrm{rec}}$, and $w_{\mathrm{ling}}$ are hyperparameters controlling the relative importance of each reward objective.

Finally, a batch-level normalization is applied:
\begin{equation}
\hat{A}^{(j)}_{\mathrm{sum}} = \frac{A^{(j)}_{\mathrm{sum}} - \mu_{\mathrm{batch}}}{\sigma_{\mathrm{batch}} + \epsilon},
\end{equation}
where $\mu_{\mathrm{batch}}$ and $\sigma_{\mathrm{batch}}$ are computed over all rollouts in the training batch.

Let $D$ denote the captioning training dataset. The corresponding multi-reward
GDPO objective is:
{\small
\begin{equation}
\mathcal{J}_{\mathrm{GDPO}}(\theta)
=
\mathbb{E}
\left[
\frac{1}{G}\sum_{j=1}^{G}\sum_{t=1}^{|o_{i,j}|}
\min\!\left(
s_{i,j,t}\hat{A}_{\mathrm{sum}}^{(i,j)},
\operatorname{clip}(s_{i,j,t},1-\epsilon,1+\epsilon)\hat{A}_{\mathrm{sum}}^{(i,j)}
\right)
\right],
\end{equation}
}
where $\mathbb{E}$ is over $(q_i,o_i)\sim D$ and
$o_{i,j}\sim\pi_{\theta_{\mathrm{old}}}(\cdot\mid q_i)$, and
$s_{i,j,t}=\pi_\theta(o_{i,j,t}\mid q_i,o_{i,j,<t})/
\pi_{\theta_{\mathrm{old}}}(o_{i,j,t}\mid q_i,o_{i,j,<t})$
is the token-level importance sampling ratio. We provide more details in Appendix~\ref{sec:implementation_settings}.

\textbf{Length-Conditional Reward Masking.} Recently there has been increasing work to train reasoning models to add length constraints~\citep{liu2025dlerdoinglengthpenalty,kimiteam2026kimik25visualagentic} for token efficiency. However, length constraints in training models with captioning-based RL serve different purposes: models could produce excessively long captions, possibly containing redundant information, in an effort to increase recall, or conversely aim to maximize precision by reducing caption length, thus missing key information. Therefore the length constraint in captioning objectives cannot be one-sided (upper bound only) as commonly done in reasoning models.

In the presence of a reference caption from either a human or a strong reference captioning model, a natural choice for a length constraint is to constrain the generated caption length with respect to the reference caption.  For example, one may use the ratio between the generated caption and the reference caption as a linear length penalty. However, such a linear length penalty can limit the exploration by prematurely encouraging the model to converge its generation length to the reference caption length, which is especially amplified when the reference caption has a very different length compared to the policy model's original caption length.

To avoid restricting exploration in the early stages of training, we instead introduce length-conditional reward masking that acts as a gating mechanism. Let $\ell_{\mathrm{pred}}$ and $\ell_{\mathrm{ref}}$ denote the token lengths of the predicted and reference captions, and define the length ratio as $\rho = \ell_{\mathrm{pred}} / \ell_{\mathrm{ref}}$. The linguistic reward is then masked by
$\tilde{R}_{\mathrm{ling}} = \begin{cases}
R_{\mathrm{ling}}, & \text{if } \tau_l \leq \rho \leq \tau_u,\\
0, & \text{otherwise}.
\end{cases}$

\section{Results}\label{sec:results}

\subsection{Experiments}

As discussed in Section~\ref{sec:intro}, considering only one aspect of captioning risks introducing bias. We use DCScore~\citep{ye2025paintingwordselevatingdetailed} to represent the correctness-and-completeness view, CaptionQA~\citep{yang2025captionqacaptionusefulimage} to represent the utility view and CapArena~\citep{cheng2025caparenabenchmarkinganalyzingdetailed} to represent the arena view. We also report average caption length on CapArena as an additional indicator of model behavior.  Additionally, we introduce b-CapScore, a balanced captioning metric that takes harmonic mean of pointability-aware precision,
reference coverage, and linguistic quality; its definition and human-alignment analysis are in
Appendix~\ref{sec:r_capscore}.

In the main results, we test our method with LLaVA1.5-7B, and QwenVL2.5 series of 3B and 7B model sizes. We compare our model with the base QwenVL2.5 models and three recent captioning-RL methods, FEEDQUILL~\citep{ye2025paintingwordselevatingdetailed}, CapRL~\citep{xing2025caprlstimulatingdenseimage} and RubiCap~\citep{huang2026rubicaprubricguidedreinforcementlearning}. We report FEEDQUILL numbers for LLaVA1.5-7B from their paper as the checkpoint is not released. CapRL is evaluated at 3B size as only 3B size is available. For RubiCap, we evaluate both the 3B and 7B checkpoints. 

Next, we perform leave-one-out ablations to assess the contribution of each component, through which we identify the causes of specific biased behaviors. Additionally, we include ablation studies to investigate the impact of reward weight (Appendix~\ref{sec:reawrd_weight_sec}), impact of using different training-time MLLM judges (Appendix~\ref{sec:robust_judge}) and additional qualitative examples (Appendix~\ref{sec:additional_qualitative}). 


\subsection{Main results}

\begin{table}[t]
  \centering
  \scriptsize
  \setlength{\tabcolsep}{3pt}
  \caption{Main results.}
  \label{tab:main_results}
  \begin{tabular}{l|cc|cc|cc|c|cc}
  \toprule

  \textbf{Model} &
  \multicolumn{2}{c|}{\textbf{DCScore $\uparrow$}} &
  \multicolumn{2}{c|}{\textbf{CaptionQA $\uparrow$}} &
  \multicolumn{2}{c|}{\textbf{CapArena $\uparrow$}} &
  \textbf{Arena Length} &
  \multicolumn{2}{c}{\textbf{b-CapScore $\uparrow$}} \\
  \midrule

 \multicolumn{10}{l}{\textit{Proprietary MLLMs}} \\
  \cmidrule(lr){1-10}
  GPT-4o           & 45.8 & -- & 78.7 & -- & 10.3 & -- &  83 & 56.1 & -- \\
  Gemini-3.1-Flash & 59.9 & -- & 83.9 & -- & 81.2 & -- & 197 & 65.6 & -- \\
  GPT-5.4          & 66.5 & -- & 89.2 & -- & 82.2 & -- & 220 & 70.9 & -- \\  
  Gemini-3.1-Pro   & 67.8 & -- & 90.0 & -- & 79.8 & -- & 362 & 73.0 & -- \\  

  \midrule

  \multicolumn{10}{l}{\textit{LLaVA-1.5-7B}} \\
  \cmidrule(lr){1-10}
  Baseline & 23.0 & -- & 46.4 & -- & -94.0 & -- & 74 & 26.9 & -- \\
  FEEDQUILL & 34.5 & \textcolor{dgreen}{+11.5} & - & - & - & - & - & -- & -- \\
  \textbf{Ours} & \textbf{36.6} & \textcolor{dgreen}{+13.6} & \textbf{55.4} & \textcolor{dgreen}{+9.0} & \textbf{-65.0} & \textcolor{dgreen}{+29.0} & 124 & \textbf{43.4} & \textcolor{dgreen}{+16.5} \\

  \midrule
  \multicolumn{10}{l}{\textit{QwenVL2.5-3B}} \\
  \cmidrule(lr){1-10}
  Baseline & 43.3 & -- & 70.0 & -- & -34.0 & -- & 131 & 46.2 & -- \\
  CapRL-3B & 48.6 & \textcolor{dgreen}{+5.3} & \textbf{82.6} & \textcolor{dgreen}{+12.6} & -50.6 & \textcolor{dred}{-16.6} & 403 & 47.4 & \textcolor{dgreen}{+1.2} \\
  RubiCap-3B & 43.0 & \textcolor{dred}{-0.3} & 71.1 & \textcolor{dgreen}{+1.1} & -29.5 & \textcolor{dgreen}{+4.5} & 149 & 46.8 & \textcolor{dgreen}{+0.6} \\
  \textbf{Ours} & \textbf{50.8} & \textcolor{dgreen}{+7.5} & 75.0 & \textcolor{dgreen}{+5.0} & \textbf{-3.8} & \textcolor{dgreen}{+30.2} & 175 & \textbf{49.3} & \textcolor{dgreen}{+3.1} \\

  \midrule
  \multicolumn{10}{l}{\textit{QwenVL2.5-7B}} \\
  \cmidrule(lr){1-10}
  Baseline & 46.0 & -- & 74.9 & -- & 13.7 & -- & 136 & 51.1 & -- \\
  RubiCap-7B & 50.5 & \textcolor{dgreen}{+4.5} & 76.0 & \textcolor{dgreen}{+1.1} & 22.7 & \textcolor{dgreen}{+9.0} & 176 & 53.9 & \textcolor{dgreen}{+2.8} \\
  \textbf{Ours} & \textbf{53.4} & \textcolor{dgreen}{+7.4} & \textbf{79.1} & \textcolor{dgreen}{+4.2} & \textbf{28.5} & \textcolor{dgreen}{+14.8} & 192 & \textbf{58.7} & \textcolor{dgreen}{+7.6} \\
  \bottomrule
  \end{tabular}
\end{table}

\begin{table}[t]
\centering
\scriptsize
\setlength{\tabcolsep}{4pt}
\caption{Performance comparison across vision benchmarks.}
\label{tab:benchmark_results}
\renewcommand{\arraystretch}{1.05}
\begin{tabular}{lccccccccccc}
\toprule
\textbf{Model} &
\rotatebox{60}{\textbf{BLINK}} &
\rotatebox{60}{\textbf{ChartQA}} &
\rotatebox{60}{\textbf{DocVQA}} &
\rotatebox{60}{\textbf{InfoVQA}} &
\rotatebox{60}{\textbf{MMBench}} &
\rotatebox{60}{\textbf{MMStar}} &
\rotatebox{60}{\textbf{OCRBench}} &
\rotatebox{60}{\textbf{ScienceQA}} &
\rotatebox{60}{\textbf{SEEDBench}} &
\rotatebox{60}{\textbf{TextVQA}} &
\rotatebox{60}{\textbf{Avg.}} \\
\midrule
QwenVL2.5-3B        & 47.91 & 83.28 & \textbf{92.85} & \textbf{74.22} & 55.88 & 55.79 & 78.70 & 83.35 & 74.94 & \textbf{78.62} & 72.55 \\
ShareGPT5V-mini-SFT & 45.11 & 82.60 & 85.92 & 69.94 & 53.92 & 56.19 & 79.80 & \textbf{83.47} & 74.51 & 66.87 & 69.83 \\
RubiCap-3B          & 45.38 & 82.84 & 90.15 & 71.26 & 51.96 & 56.29 & 75.90 & 82.60 & \textbf{75.77} & 75.13 & 70.73 \\
CapRL-3B            & 47.88 & 83.40 & 90.18 & 73.17 & \textbf{57.84} & \textbf{56.94} & \textbf{81.00} & 83.40 & 75.67 & 73.26 & 72.27 \\
BalCapRL-3B         & \textbf{47.92} & \textbf{83.64} & 92.78 & 74.17 & 56.86 & 55.93 & 78.70 & 83.31 & 75.55 & 78.40 & \textbf{72.73} \\
\bottomrule
\end{tabular}

\end{table}


\textbf{BalCapRL consistently outperforms prior work in  captioning benchmarks.} As shown in Table~\ref{tab:main_results}, with LLaVA-1.5-7B, our method strongly improves over the baseline across all metrics, lifting DCScore by 13.6 points, CaptionQA by 9.0 points, and CapArena by 29.0 points. Compared to FEEDQUILL, which arguably optimizes DCScore, our method still outperforms it by 2.1 points on this metric.

Using QwenVL2.5-3B as the base model, compared to CapRL-3B, our method strongly outperforms it in DCScore and CapArena, though CapRL-3B scores higher still in CaptionQA. Notably, CapRL-3B produces captions roughly 3$\times$ longer than the base policy, and even regresses compared to it in CapArena by 16.6 points. We believe this to be a direct result of CapRL's optimization method, which strictly optimizes captions for MQA utility, leading to excessively long captions with degraded fluency as also demonstrated in the qualitative examples in Appendix~\ref{sec:additional_qualitative}. In comparison, our method's balanced objective improves over the baseline across all metrics.

Compared to RubiCap-3B, our method strongly outperforms it on all metrics, even approaching the larger RubiCap-7B in DCSCore and CaptionQA performance. When using the same QwenVL2.5-7B base model, our method again significantly outperforms RubiCap-7B on all evaluated benchmarks.

\textbf{BalCapRL largely preserves general vision benchmark performance.} Beyond captioning benchmarks, we also study the performance of models on ten general vision benchmarks in Table~\ref{tab:benchmark_results}. We first create a baseline that finetunes the base model using Supervised Fine-tuning (SFT) on the same RL training data (referred as ShareGPT5V-mini) and then evaluate captioning-RL models. The results show that while SFT improves the performance in some benchmarks such as ScienceQA, it loses nontrivial performance in most benchmarks compared to its base model, confirming the findings of RubiCap~\citep{huang2026rubicaprubricguidedreinforcementlearning}. Surprisingly, while RL is commonly believed to suffer less from catastrophic forgetting, prior work such as RubiCap and CapRL models still suffer from the regression, potentially due to their imbalanced reward design. We note that CapRL-3B significantly improves performance in MMBench and OCRBench while suffering from nontrivial degradation in TextVQA and DocVQA. In contrast, BalCapRL has no notable regression in any tested benchmark while having improvements on several benchmarks.

\textbf{The proposed method is robust to the tested judge choices.} We show in Appendix~\ref{sec:robust_judge} the method remains effective when varying the choice of the judge models (i.e., GPT-4o-mini, GPT-5-mini, GPT-5.4). Note that results in Table~\ref{tab:main_results} were obtained via GPT-4o-mini judge for the low cost and fast training, while stronger judge such as GPT-5.4 could yield even better results.

\subsection{Ablation Studies}

\begin{table}[htbp]
\centering
\scriptsize
\caption{Leave-one-out ablation studies. $w_{\mathrm{prec}}$, $w_{\mathrm{rec}}$, $w_{\mathrm{ling}}$ are set equal for this study.}
\label{tab:leave_one_ablation}
\begin{tabular}{lcccc}
\toprule
\textbf{Model} & \textbf{DCScore} & \textbf{CaptionQA} & \textbf{CapArena} & \textbf{Arena Length} \\
\midrule
QwenVL2.5-3B      & 43.3 & 70.0  & -34.0  & 131 \\
Full            & 52.0 & 75.0  & -12.0  & 203 \\
w/o c-GDPO        & 38.0 & 67.0  & -71.8  & 96  \\
w/o Precision (CapArena-biased)  & 41.2 & 73.6  & -13.8  & 163 \\
w/o Recall      & 51.8 & 74.9  & -19.3  & 152 \\
w/o Linguistic (Utility-biased)  & 53.4 & 76.3  & -51.0  & 375 \\
w/o Pointability     & 39.2 & 63.5  & -85.7  & 240 \\
w/o recap with gpt-5-mini    & 46.8 & 73.6  & -17.3  & 152 \\


\bottomrule
\end{tabular}
\end{table}

\begin{table}[htbp]
\centering
\scriptsize
\caption{Ablation studies over choice of length penalty and hyperparameters.}
\label{tab:length_ablation}
\begin{tabular}{lcccc}
\toprule
\textbf{Model} & \textbf{DCScore} & \textbf{CaptionQA} & \textbf{CapArena} & \textbf{Arena Length} \\
\midrule
QwenVL2.5-3B           & 43.3 & 70.0  & -34.0  & 131 \\
w/o length penalty     & 39.0 & 71.6  & -32.5  & 81 \\
w/ linear length penalty & 52.7 & 72.7  & -33.0  & 90 \\
w/ proposed length penalty & 52.0 & 75.3 & -19.0 & 203 \\
\midrule
\multicolumn{5}{l}{\textit{Length-Conditional Reward Masking ($\tau_l$, $\tau_u$)}} \\
\midrule
$\tau_l=0$, $\tau_u=3$     & 39.4  & 71.4 & -31.0  & 80 \\
$\tau_l=0.5$, $\tau_u=1$   & 46.1  & 74.0 & -17.7  & 121 \\
$\tau_l=0.5$, $\tau_u=2$   & 52.0  & 75.3 & -19.0  & 203 \\
$\tau_l=0.5$, $\tau_u=3$   & 54.0  & 75.9 & -62.7  & 313 \\
$\tau_l=0.5$, $\tau_u=4$   & 55.4  & 75.9 & -54.0  & 322 \\
$\tau_l=0.5$, $\tau_u=5$   & 54.8  & 75.5 & -72.0  & 355 \\
$\tau_l=0.5$, $\tau_u=6$   & 54.5  & 75.8 & -36.7  & 280 \\
\bottomrule
\end{tabular}
\end{table}

To assess the impact of each component of our method, we start with leave-one-out ablation studies in Table~\ref{tab:leave_one_ablation}, followed by an ablation study on length penalty in Table~\ref{tab:length_ablation}. We focus on QwenVL2.5-3B as it is the most commonly used model among prior captioning-RL work. 

\textbf{Keeping c-GDPO is critical in our setting.} We first examine the effect of removing c-GDPO. Instead of applying separate group normalization to each reward, as in c-GDPO, we follow vanilla GRPO by summing the three rewards and applying group normalization only to the aggregated reward. Relative to the QwenVL2.5-3B baseline, vanilla GRPO leads to substantial performance degradation across benchmarks. We attribute this to the vanilla GRPO's difficulty in learning fine-grained signals of multiple rewards with distinct dynamics (Figure~\ref{fig:gdpo_illustration}).

\textbf{Effects of removing individual rewards.} We then perform an ablation study by removing each of the three rewards from the full method (Table~\ref{tab:leave_one_ablation}), setting the corresponding reward weight to zero. Removing the precision reward maintains the large gain in CapArena but leads to a clear drop in DCScore. We denote this variant as the CapArena-biased model in Figure~\ref{fig:Fig1}. This behavior is expected: without the precision constraint, the model can more freely optimize toward matching the reference captions (generated by GPT-5-mini) and improving linguistic score, even when doing so reduces visually-verifiable precision and harms DCScore.
In contrast, removing the recall reward yields performance above the baseline on all benchmarks and remains only slightly below the full method. This suggests that our framework still improves the model without relying on the recall reward. 
Notably, removing the linguistic reward increases both CaptionQA and DCScore compared to even our full method, but causes a substantial drop in CapArena. We label this variant, which somewhat resembles CapRL in behavior, as the utility-biased model in Figure~\ref{fig:Fig1}. Similar to CapRL-3B, we observe an approximately 3$\times$ increase in caption length, suggesting that the model may generate repetitive or overly enumerative content at the expense of fluency and coherence. These results highlight that linguistic quality is not adequately captured by CaptionQA and DCScore alone, and should therefore be explicitly considered in captioning RL.

\begin{figure}[h]
    \centering
    \includegraphics[width=\textwidth]{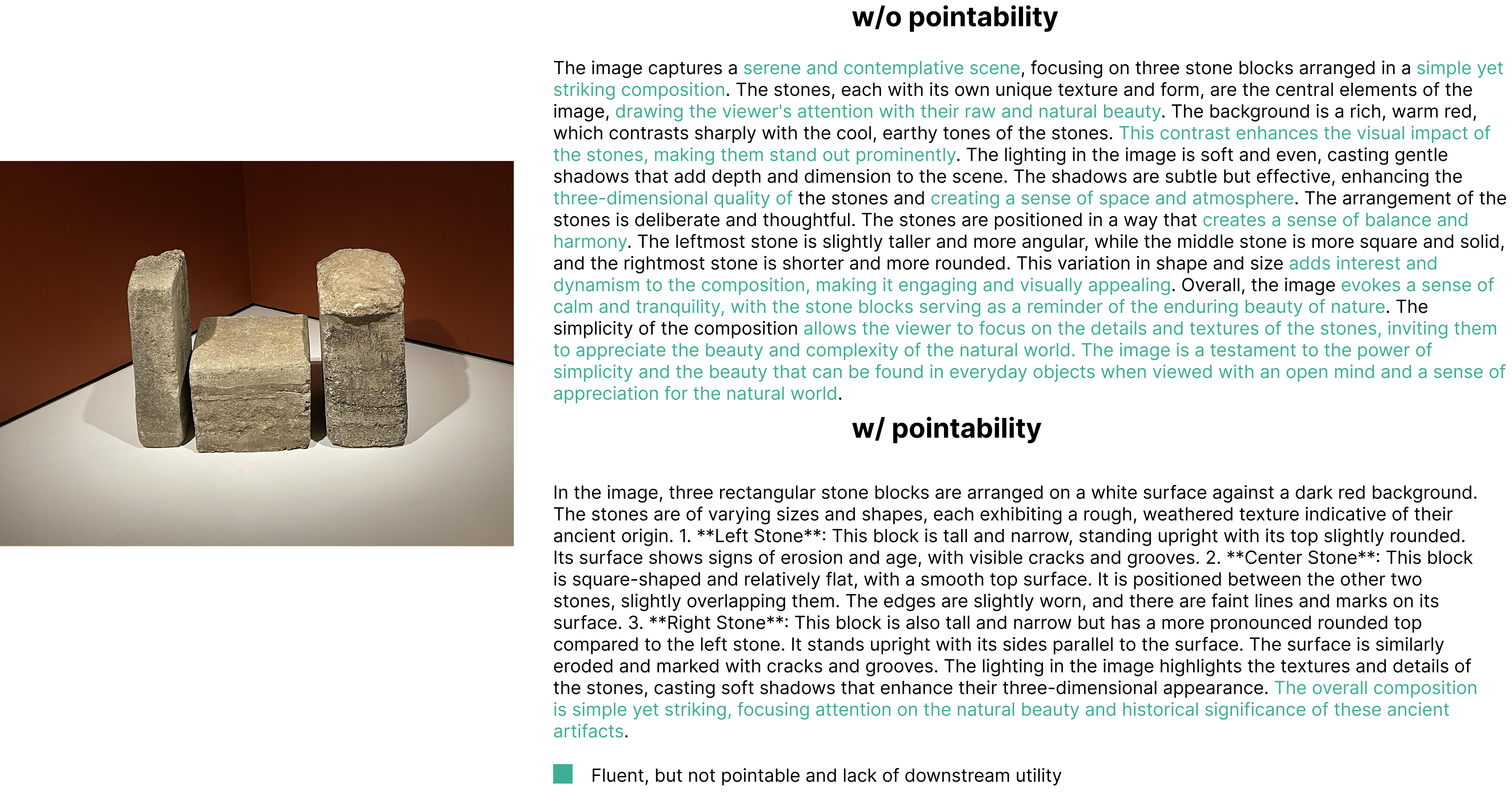}
\caption{\textbf{A qualitative comparison between with pointability rubric and without.}}
    \label{fig:pointability_qualitative_example}
\end{figure}

\textbf{Keeping the pointability rubric mitigates meta commentary.} Removing the pointability rubric largely hurts the model (Table~\ref{tab:leave_one_ablation}). A qualitative examination in Figure~\ref{fig:pointability_qualitative_example} shows that without pointability rubric, the model learns to hack the rewards by abusing the use of meta-commentary (fluent but lacking downstream utility).

\textbf{Better reference caption helps.} We examine the impact of reference-caption quality by replacing the GPT-5-mini recapped captions with the original ShareGPT4V captions (see Table~\ref{tab:leave_one_ablation}). This results in consistent drops across benchmarks, suggesting that higher-quality reference captions provide a more effective recall signal under our framework. Notably, this variant performs slightly worse than removing the recall reward altogether. This suggests that a weak or poorly aligned reference-caption signal may be less beneficial than omitting the recall objective entirely, under our framework.

\textbf{Length penalty ablation studies.} We study the impact of the length penalty in Table~\ref{tab:length_ablation}. We begin by removing the length penalty from our framework. Under this setting, the model falls into a suboptimal regime in which it regresses in DCScore and slightly improves in CaptionQA and CapArena. Note that its caption length becomes shorter than its base model, potentially exploiting shorter captions to avoid hallucinations too aggressively. This is undesirable as we expect to have the model generate more detailed captions with the same or fewer hallucinations. We then add a length penalty that penalizes deviations of the predicted-to-reference caption length ratio from a predefined acceptable interval in a piecewise linear manner (Appendix~\ref{sec:linear_length}). This greatly improves DCScore, suggesting the importance of length penalty for captioning-RL. Then we show that our length-conditional reward masking outperforms the linear length penalty in a more balanced way and resulting in longer caption length. Next we study our length-conditional reward masking by varying $\tau_l$ and $\tau_u$. We observed that setting an upper bound $\tau_u$ is not sufficient to prevent the model from generating short captions to over-avoid mistakes. In contrast, setting $\tau_l$ to 0.5 effectively mitigates this behavior. Next we fix $\tau_l$ to 0.5 and study the effect of varying $\tau_u$ from 1 to 6. We observe that allowing the model to generate longer captions would generally improve CaptionQA and DCScore, but at the expense of CapArena performance. 

\section{Related work}\label{sec:related_work}

\textbf{Reinforcement learning for MLLMs on image captioning.} Reinforcement learning has been used in MLLMs to mitigate hallucinations~\citep{sun2023aligninglargemultimodalmodels}, improve multimodal reasoning~\citep{feng2025videor1reinforcingvideoreasoning}, and better align model outputs with human preferences~\citep{yu2025rlaifvopensourceaifeedback}. More generally, reinforcement learning with verifiable rewards (RLVR)~\citep{shao2024deepseekmathpushinglimitsmathematical} is effective when the target outcome can be checked automatically. For tasks without simple verifiable outcomes, prior works often rely on LLM/MLLM feedback signals~\citep{lee2024rlaifvsrlhfscaling,gunjal2025rubricsrewardsreinforcementlearning, yu2025rlaifvopensourceaifeedback}. Recently, captioning-centric RL work~\citep{ye2025paintingwordselevatingdetailed,xing2025caprlstimulatingdenseimage,huang2026rubicaprubricguidedreinforcementlearning, zhang2025sccaptionerimprovingimagecaptioning,tang2026cccaptiondualrewardreinforcementlearning} has gained increasing attention. 
FEEDQUILL~\citep{ye2025paintingwordselevatingdetailed} decomposes generated captions into atomic assertions and uses MLLMs to assess whether each assertion is visually verifiable or covered by the reference captions. Similar to FEEDQUILL, our method also incorporates both precision and recall into the reward. However, unlike FEEDQUILL, which requires two separate reward models, our approach directly uses MLLM-judge which greatly simplifies the pipeline while introducing a latency and cost tradeoff. In addition, we include a pointability rubric to the precision reward and include linguistic reward to discourage repetitive content that degrades fluency and coherence.
CapRL~\citep{xing2025caprlstimulatingdenseimage} uses a two-stage pipeline in which generated captions are sent to a text-only LLM for question answering, and answer accuracy is used as the reward. In contrast, we avoid explicit MQA sampling and instead encourage caption utility through a simple notion of pointability. This provides finer-grained supervision at the caption level and reduces the risk of over-optimizing for downstream MQA performance alone.
\mbox{RubiCap~\citep{huang2026rubicaprubricguidedreinforcementlearning}} uses a committee of strong MLLMs to construct per-sample rubrics for rubric-based reinforcement learning. In contrast, our method does not require multiple MLLMs to generate sample-specific rubrics. We also explicitly incorporate a utility-oriented reward component, which leads to improved caption utility compared to RubiCap.

\textbf{Image Captioning metrics.} Traditional image captioning metrics, including BLEU~\citep{papineni2002bleu}, METEOR~\citep{banerjee-lavie-2005-meteor}, and CIDEr~\citep{vedantam2015ciderconsensusbasedimagedescription}, measure similarity between generated captions and human references using n-gram overlap. While widely used, such metrics are sensitive to surface-level phrasing variations and therefore struggle to capture the diversity of valid descriptions. Model-based metrics, such as SPICE~\citep{anderson2016spicesemanticpropositionalimage}, CAPTURE~\citep{pothiraj2025captureevaluatingspatialreasoning}, and CLIPScore~\citep{hessel2022clipscorereferencefreeevaluationmetric}, were introduced to alleviate this limitation. However, recent studies show that these metrics become less reliable for the long, detailed, and nuanced captions produced by modern MLLMs in open-ended captioning settings.
To evaluate MLLM captions, recent work increasingly relies on LLMs or strong commercial MLLMs as judges. These approaches can be broadly categorized into three perspectives: (1) \textbf{the utility view}, which measures how well a caption supports downstream text-only question answering, as in Prism~\citep{qiao2024prismframeworkdecouplingassessing} and CaptionQA~\citep{yang2025captionqacaptionusefulimage}; (2) \textbf{the correctness-and-completeness view}, which evaluates faithfulness and coverage of the captions ~\citep{ye2025paintingwordselevatingdetailed,jing2024faithscorefinegrainedevaluationshallucinations,liu2025capabilitycomprehensivevisualcaption}; and (3) \textbf{the arena view}, which assesses caption quality through pairwise competition, as in CapArena~\citep{cheng2025caparenabenchmarkinganalyzingdetailed}. Our work argues that these aspects should be combined rather than considered separately, thereby reducing the biases introduced by any single evaluation aspect.

\section{Limitations}\label{sec:limitations}

One limitation of our work is that some aspects of caption quality are not explicitly modeled in our reward design. In particular, plausible inferences supported by world knowledge may be under-rewarded by the pointability rubric, which favors visually pointable and directly verifiable content. Such information can only be indirectly preserved through the recall objective, making performance on these aspects dependent on the quality and coverage of the reference captions. As a result, our framework may undervalue captions that contain reasonable but non-pointable inferences. Another limitation is our method uses MLLM-as-judge, which largely simplifies the pipeline compared to related method FEEDQUILL but introduces latency and cost tradeoff.

\bibliographystyle{unsrtnat} 
\bibliography{references}
\appendix
\section{Appendix}

\subsection{Implementation Details}\label{sec:implementation_settings}

We build our codebase and implement c-GDPO on top of verl~\cite{sheng2024hybridflow}. We set the number of rollouts to 8, the learning rate to 5e-6, and the batch size to 256. We use a cosine learning rate scheduler and train for 1 epoch. For main results in Table~\ref{tab:main_results}, we use $w_{\mathrm{pre}}$ = 0.1, $w_{\mathrm{rec}}$ = 0.3, $w_{\mathrm{ling}}$ = 0.3, $\tau_l$ = 0.5 , and $\tau_u$ = 2 for better empirical results.

Original  ShareGPT4V~\cite{chen2023sharegpt4vimprovinglargemultimodal} dataset contains roughly 90K image-text pairs and its captions are annotated by GPT-4V~\cite{openai2024gpt4technicalreport}.  We compare the original captions with a recapped version generated by GPT-5-mini. In Table~\ref{tab:main_results}, GPT-4o-mini is used to decompose on-policy captions into atomic assertions and to label each assertion and also used to label assertions for whether it is correct, pointable and covered in the reference caption. 

We empirically observed that removing the KL divergence loss~\cite{yu2025dapoopensourcellmreinforcement} and applying dual-clip~\cite{ye2020masteringcomplexcontrolmoba} improves training stability (which is an implementation detail omitted from Section~\ref{sec:method} to keep it simple). Following DAPO~\cite{yu2025dapoopensourcellmreinforcement}, we also adopted ``\texttt{token-sum-sequence-mean}'' which further improves empirical performance.

Every single experiment is conducted using one 8-gpus B-200 node. For the main results that use GPT-4o-mini as judge, it takes roughly 24 hours for LLaVA-7B experiments, 37 hours for QwenVL2.5-3B experiments and 46 hours for QwenVL2.5-7B experiments.

\subsection{Training Prompt}
\label{sec:training_prompt}

\begin{lstlisting}
You are an expert image caption evaluator. You will analyze a SYNTHETIC CAPTION and a GROUND TRUTH CAPTION for an image to extract detailed features for quality assessment.

SYNTHETIC CAPTION (to evaluate):
"%(caption)s"

GROUND TRUTH CAPTION (reference):
"%(g_caption)s"

Your task is to extract the following features in JSON format:

1. Atomic Assertions: Break each caption into shortest possible factual claims.
   Extract ALL assertions from the caption, including meta-commentary and subjective statements.
   Do NOT filter out any assertions during extraction - we need the complete list.

2. Precision and Recall - VERIFICATION REQUIRES BOTH CONDITIONS:

   =======================================================================
   VERIFICATION RUBRIC: TWO-PART TEST
   =======================================================================

   For EACH synthetic assertion, mark is_verified: true ONLY if BOTH conditions are met:

   CONDITION 1 - THE "POINT TO IT" TEST:
   Could a person physically POINT TO this thing in the image?

   [PASS] Physical objects, visible attributes, spatial locations, visible text
     - "a car", "a red door", "silver hubcap", "on the left", "sign reads STOP"

   [FAIL] Abstractions, effects, judgments, emotions, relationships, intentions
     - "adds depth", "contrasting nicely", "inviting the viewer", "happy", "family"

   CONDITION 2 - VISUAL VERIFICATION:
   Is this assertion actually TRUE in the image?
   Check against the ground truth caption OR examine the image directly.

   [PASS] The described object/attribute/location is actually visible in the image
   [FAIL] The assertion is a hallucination or factually incorrect

   =======================================================================

   VERIFICATION DECISION TABLE:

   | Assertion            | Pointable? | Actually in Image? | is_verified |
   |----------------------|------------|--------------------|-------------|
   | "A silver hubcap"    | Yes        | Yes (exists)       | TRUE        |
   | "A silver hubcap"    | Yes        | No (no hubcap)     | FALSE       |
   | "adds depth"         | No         | N/A                | FALSE       |
   | "contrasting nicely" | No         | N/A                | FALSE       |
   | "A purple elephant"  | Yes        | No (not there)     | FALSE       |

   KEY PRINCIPLE: An assertion that fails EITHER test must be marked is_verified: false.
   - Meta commentary fails the "point to it" test -> is_verified: false
   - Hallucinated objects fail the visual verification test -> is_verified: false
   - Only correct, pointable facts pass both tests -> is_verified: true

   For RECALL: Check if synthetic caption covers GT content (exact wording not needed)

3. Evaluate the linguistic quality of the synthetic caption, independent of its accuracy to the image. Assess the following:
- Clarity: How easily can a human reader understand the caption? Consider absence of ambiguous or overly complex phrasing, simplicity and readability, avoidance of unnecessary detail that obscures meaning. Penalize overly long or overloaded sentences that pack too many ideas or modifiers into a single sentence, making the caption difficult to parse or unnatural for human readers. Penalize unnecessary repetition or redundant wording that does not add new information.
- Fluency: How natural and well-formed is the writing? Consider correct grammar, punctuation, and spelling, natural sentence flow, human-like phrasing, not a list of facts, avoidance of unnatural compound words, excessive adjective stacking, or run-on sentences.
- Coherency: How well do different parts of the caption connect to form one unified statement? Consider logical ordering of information, smooth transitions between ideas, consistent perspective and focus, no abrupt topic shifts.

Based on the above criteria, rate each item (Clarity, Fluency, Coherency) from 1 to 10, where 10 is the highest score given for excellent, highly natural, and easy to read synthetic caption and 1 is the lowest score for very poor, confusing, or unnatural synthetic caption. After rating, provide a brief summary explaining the reasoning behind the given quality scores especially noting problems such as convoluted syntax, excessive modifiers, disjointed structure, unnatural or robotic phrasing, invented or overly technical compound words.

Return ONLY valid JSON in this EXACT format (no markdown, no code blocks):
{
    "synthetic_features": {
        "atomic_assertions": [
            {"text": str, "is_verified": bool},
            ...
        ],
        "clarity_score": int,
        "fluency_score": int,
        "coherency_score": int,
        "linguistic_scores_explanation": str
    },
    "gt_features": {
        "atomic_assertions": [
            {"text": str, "is_covered": bool},
            ...
        ]
    }
}

IMPORTANT RULES:
- EXTRACT ALL: Include ALL assertions from the caption, even meta-commentary and subjective statements.
- TWO-PART TEST: is_verified: true requires BOTH (1) pointable AND (2) actually visible in image.
- PENALIZE BOTH: Meta commentary AND hallucinations should both result in is_verified: false.
- Be strict: when in doubt about either condition, mark as NOT VERIFIED.
- For recall: check if synthetic caption covers GT content (not exact wording needed)
- Return ONLY the JSON object, no other text or formatting
\end{lstlisting}

\subsection{Full Proof}
\label{sec:full_proof}

\paragraph{Proof of Proposition 1.}
Consider a $K$-reward, $G$-rollout setting. We assume G >= 3 to ensure the group-wise standard deviation is well-defined and the resulting normalization is non-trivial.  Let the reward vector of rollout $i$ be
\[
\mathbf{r}_i = (r_{i,1},\dots,r_{i,K}) \in \mathbb{R}^K,
\]
and let $w_k \in \mathbb{R}$ denote the aggregation weight for reward dimension $k$.

\subparagraph{Vanilla GRPO.}
Vanilla multi-reward GRPO first aggregates rewards into a single scalar
\[
s_i = \sum_{k=1}^K w_k r_{i,k}.
\]
It then computes the group-normalized advantage from the scalar scores $\{s_i\}_{i=1}^G$:
\[
A_i^{\mathrm{GRPO}}
=
\frac{s_i-\mu_s}{\sigma_s},
\qquad
\mu_s = \frac{1}{G}\sum_{j=1}^G s_j,
\qquad
\sigma_s = \sqrt{\frac{1}{G}\sum_{j=1}^G (s_j-\mu_s)^2},
\]
where we use the population standard deviation for concreteness. The same argument holds under sample normalization up to a constant factor.

Fix rollout $i=1$, and fix the competing rollouts $\mathbf{r}_2,\dots,\mathbf{r}_G$. Then $s_2,\dots,s_G$ are fixed constants, and the only variable is $\mathbf{r}_1$. Since
\[
s_1 = \sum_{k=1}^K w_k r_{1,k},
\]
both $\mu_s$ and $\sigma_s$ are functions of $s_1$ only:
\[
\mu_s = \frac{s_1 + \sum_{j=2}^G s_j}{G},
\qquad
\sigma_s = \sigma_s(s_1; s_2,\dots,s_G).
\]
Therefore there exists a scalar function $f:\mathbb{R}\to\mathbb{R}$, depending on the fixed competing rollouts, such that
\[
A_1^{\mathrm{GRPO}} = f(s_1) = f\!\left(\sum_{k=1}^K w_k r_{1,k}\right).
\]
Hence the normalized advantage of rollout $1$ depends on its reward vector $\mathbf{r}_1$ only through the scalar weighted sum $\sum_{k=1}^K w_k r_{1,k}$.

Now consider two reward vectors $\mathbf{r}_1,\mathbf{r}_1' \in \mathbb{R}^K$ such that
\[
\sum_{k=1}^K w_k r_{1,k}
=
\sum_{k=1}^K w_k r'_{1,k}.
\]
Then $s_1 = s_1'$, which implies
\[
A_1^{\mathrm{GRPO}}(\mathbf{r}_1)
=
f(s_1)
=
f(s_1')
=
A_1^{\mathrm{GRPO}}(\mathbf{r}_1').
\]
Therefore, all reward vectors lying on the same weighted-sum hyperplane
\[
\left\{
\mathbf{r}\in\mathbb{R}^K : \sum_{k=1}^K w_k r_k = c
\right\}
\]
are indistinguishable to vanilla GRPO. This proves that reward aggregation before group normalization induces a many-to-one mapping from $\mathbb{R}^K$ to scalar normalized advantages, and that optimization depends only on the aggregated reward direction while discarding orthogonal reward-trade-off information.

\subparagraph{c-GDPO.}
In reward-decoupled normalization, each reward dimension is normalized separately before aggregation. Define
\[
\mu_k = \frac{1}{G}\sum_{j=1}^G r_{j,k},
\qquad
\sigma_k = \sqrt{\frac{1}{G}\sum_{j=1}^G (r_{j,k}-\mu_k)^2},
\]
and let the normalized per-reward deviation of rollout $i$ be
\[
\widetilde{A}_{i,k} = \frac{r_{i,k}-\mu_k}{\sigma_k}.
\]
The final decoupled advantage is
\[
A_i^{\mathrm{c\text{-}GDPO}}
=
\sum_{k=1}^K w_k \widetilde{A}_{i,k}
=
\sum_{k=1}^K w_k \frac{r_{i,k}-\mu_k}{\sigma_k}.
\]

Again fix rollout $i=1$ and competing rollouts $\mathbf{r}_2,\dots,\mathbf{r}_G$. Then for each reward dimension $k$, the quantities $\mu_k$ and $\sigma_k$ depend on $r_{1,k}$ and the fixed values $r_{2,k},\dots,r_{G,k}$, but do not depend on any other coordinate $r_{1,\ell}$ with $\ell\neq k$. Hence there exist scalar functions $f_k:\mathbb{R}\to\mathbb{R}$ such that
\[
A_1^{\mathrm{c\text{-}GDPO}}
=
\sum_{k=1}^K w_k f_k(r_{1,k}).
\]
Therefore the decoupled advantage depends on the reward coordinates separately rather than only through the aggregated sum $\sum_{k=1}^K w_k r_{1,k}$.

Consequently, reward-decoupled normalization is generally not invariant to the weighted-sum hyperplanes above. In particular, suppose there exist two reward dimensions $p\neq q$ with nonzero weights $w_p,w_q$, and nondegenerate competing-rollout variances in those dimensions so that $f_p$ and $f_q$ are not constant. Consider two reward vectors $\mathbf{r}_1,\mathbf{r}_1'$ that differ only in coordinates $p$ and $q$ and satisfy
\[
w_p r_{1,p} + w_q r_{1,q}
=
w_p r'_{1,p} + w_q r'_{1,q},
\]
with $r_{1,p}\neq r'_{1,p}$ and $r_{1,q}\neq r'_{1,q}$. Then $\mathbf{r}_1$ and $\mathbf{r}_1'$ lie on the same weighted-sum hyperplane, but
\[
A_1^{\mathrm{c\text{-}GDPO}}(\mathbf{r}_1)
-
A_1^{\mathrm{c\text{-}GDPO}}(\mathbf{r}_1')
=
w_p\!\left[f_p(r_{1,p})-f_p(r'_{1,p})\right]
+
w_q\!\left[f_q(r_{1,q})-f_q(r'_{1,q})\right],
\]
which is generically nonzero because the two coordinates are normalized independently. Thus equal aggregate reward does not in general imply equal decoupled advantage.

Therefore, unlike vanilla GRPO, reward-decoupled normalization does not collapse all reward trade-offs sharing the same aggregated reward into the same update signal. Instead, it preserves per-reward relative deviations before aggregation, retaining finer-grained optimization information for continuous multi-reward trade-offs.
\hfill 

\subsection{Evaluation Prompt (b-CapScore)}
\label{sec:evaluation_prompt}

Same to the training prompt, but slight differences in the precision verification part to allow generalization. More specifically:

\begin{lstlisting}
 
=======================================================================
VERIFICATION RUBRIC: TWO-PART TEST
=======================================================================

For EACH synthetic assertion, mark is_verified: true ONLY if BOTH conditions are met:

CONDITION 1 - THE "POINT TO IT OR POINT TO EVIDENCE THAT JUSTIFIES IT" TEST:
Could a person either:
(1) physically POINT TO the asserted thing in the image, OR
(2) point to visible evidence that reasonably JUSTIFIES the assertion using ordinary world knowledge?

 PASSES - Directly Visible Facts:
  - "a car"
  - "a red door"
  - "silver hubcap"
  - "on the left"
  - "sign reads STOP"

 PASSES - Visually Justified Inferences Using World Knowledge:
  - "the road is wet": puddles, reflections, surface sheen
  - "the building is under construction": scaffolding, unfinished surfaces, exposed structure
  - "the person is a bride": wedding dress, veil, bouquet, ceremony context
  - "the man is likely a chef": chef hat, apron, kitchen setting
  - "the child is blowing out birthday candles": cake, candles, leaning posture
  - "the people are waiting in line": queue formation toward a counter or entrance
  - "the room is set up for a meeting": conference table, arranged chairs, screen, notebooks
  - "the person is reading a menu": menu-like object, restaurant setting, gaze/posture

These pass because the claim is not directly visible as a single object, but the image provides clear evidence that justifies the inference using ordinary shared world knowledge.

 FAILS - Unsupported Speculation:
  - "adds depth to the composition"
  - "contrasting nicely"
  - "inviting the viewer"
  - "creates a welcoming atmosphere"
  - "the photographer intended to..."
  - "the person feels proud"
  - "this suggests luxury" (without very strong visible evidence)
  - "they are a family" (unless the image gives unusually strong evidence)
  - "this is a reunion"
  - "the person is thinking about leaving"

IMPORTANT:
World knowledge IS allowed when it is used to make a normal, evidence-based visual inference.
World knowledge is NOT allowed when it becomes mind-reading, storytelling, symbolism, or weak social speculation.

CONDITION 2 - VISUAL VERIFICATION:
Is this assertion actually TRUE in the image?

Use the image as the PRIMARY source of truth.
The ground-truth caption may help, but do not reject a claim only because it is not mentioned in the ground-truth caption.

 PASSES:
  - the assertion is visually supported and actually true

 FAILS:
  - the assertion is hallucinated
  - the assertion is factually incorrect
  - the evidence is too weak or ambiguous to support the inference confidently

=======================================================================

KEY PRINCIPLE:
An assertion should be marked is_verified: true only if it is either directly visible OR justified by visible evidence through ordinary world knowledge, AND it is actually true for this image.

\end{lstlisting}

\subsection{A Balanced Captioning Metric (b-CapScore)}
\label{sec:r_capscore}

While we advocate to use multiple captioning benchmarks to evaluate the quality of the captioning, we show that it is possible to turn our reward to one metric to evaluate the captioning quality in a more balanced way. 

We reuse the images and human reference captions of DCScore~\cite{ye2025paintingwordselevatingdetailed} as the data sources and define the balanced metric b-CapScore as the harmonic mean of
pointability-aware precision, reference coverage, and linguistic quality:
\begin{equation}
\mathrm{b\text{-}CapScore}
=
\frac{3}
{
\frac{1}{R_{\mathrm{prec}} }
+
\frac{1}{R_{\mathrm{rec}} }
+
\frac{1}{R_{\mathrm{ling}} }
}.
\end{equation}
Here, $R_{\mathrm{prec}}$ is computed using the same pointability-aware
verification rubric as in training: an atomic assertion is counted as correct
only if it is visually verified and pointable, or supported by pointable visual
evidence (full prompt in Appendix~\ref{sec:evaluation_prompt}). 
$R_{\mathrm{rec}}$ measures reference coverage, and $R_{\mathrm{ling}}$ denotes
the normalized linguistic-quality score. By using a harmonic mean, b-CapScore
penalizes imbalance across correctness, coverage, and linguistic quality. To ensure we reduce the error from atomic assertion decomposition, we use GPT-5.4 as the judge for b-CapScore evaluation for Table~\ref{tab:main_results}.

\begin{table}[H]
\centering
\small
\caption{Comparison on Model-level Spearman}
\label{tab:metric_comparison}
\begin{tabular}{lcc}
\toprule
Metric &  Ranking Procedure & Model-level Spearman \\
\midrule
GPT-4o-as-a-Judge w/ ref
 
& Arena/ELO from pairwise judgments 
& 0.943 \\
DCScore
 
& Average score over captions
& 0.943 \\
b-CapScore 
 
& Average score over captions 
& \textbf{0.956} \\
\bottomrule
\end{tabular}

\label{tab:model_level_spearman}
\end{table}

To better understand the proposed balanced captioning metric, we perform human alignment analysis on CapArena and compare the human alignment between CapArena and b-CapScore in Table~\ref{tab:metric_comparison}. 
CapArena reports model-level Spearman by converting metric judgments into pairwise arena outcomes and deriving model rankings through the same arena/ELO-style procedure used for human preferences. 
In contrast, our metric does not require arena-style pairwise comparison: for each model, we compute the average reference-based caption score over the evaluation set, rank models by this average score, and report Spearman's rank correlation with the human-derived CapArena ranking. This shows that our reward can be turned to a metric that has comparable or better human alignment than arena-style comparison.

\subsection{Reward Weight Ablation Studies}
\label{sec:reawrd_weight_sec}

\begin{table}[H]
\centering
\caption{Results for different values of $w_{\text{pre}}$.}
\begin{tabular}{lccc}
\hline
$w_{\text{pre}}$ & DCScore & CaptionQA & CapArena \\
\hline
0.0 & 41.2 & 73.6 & -13.8 \\
0.1 & 50.8 & 75.0 & -3.8  \\
0.2 & 52.8 & 74.9 & -5.8  \\
0.3 & 52.0 & 75.0 & -12.0 \\
0.4 & 52.1 & 75.3 & -10.0 \\
\hline
\end{tabular}
\label{tab:weight_ablation}
\end{table}

In Table~\ref{tab:weight_ablation}, we analyze sensitivity to the reward weights by varying $w_{\text{prec}}$ while keeping the other two reward weights fixed. Setting $w_{\text{prec}}=0$ leads to the worst overall performance. For non-zero values of $w_{\text{prec}}$, increasing its weight introduces trade-offs across benchmarks.

\subsection{Impact of training-time MLLM judge}
\label{sec:robust_judge}

\begin{table}[H]
\centering
\caption{Comparison of using different training-time judge.}
\label{tab:judge_comparison}
\begin{tabular}{lcccc}
\toprule
\textbf{Model} & \textbf{DCScore} & \textbf{CaptionQA} & \textbf{CapArena} & \textbf{Arena Length} \\
\midrule
\multicolumn{5}{l}{\textit{3B models}} \\
QwenVL2.5-3B              & 43.3 & 70.0 & -34.0 & 131 \\
BalCapRL-3B w/ GPT-4o-mini   & 50.8 & 75.0 & \textbf{-3.8}  & 175 \\
BalCapRL-3B w/ GPT-5-mini    & \textbf{50.9} & 75.4 & -6.3  & 186 \\
BalCapRL-3B w/ GPT-5.4       & 50.8 & \textbf{75.9} & -5.2  & 163 \\
\midrule
\multicolumn{5}{l}{\textit{7B models}} \\
QwenVL2.5-7B              & 46.0 & 74.9 & 13.7  & 136 \\
BalCapRL-7B w/ GPT-4o-mini   & 53.4 & 79.1 & 28.5  & 192 \\
BalCapRL-7B w/ GPT-5-mini    & 52.2 & 80.8 & 35.3  & 191 \\
BalCapRL-7B w/ GPT-5.4       & \textbf{54.2} & \textbf{81.1} & \textbf{41.7} & 172 \\
\bottomrule
\end{tabular}
\end{table}

\subsection{Additional Qualitative Examples}
\label{sec:additional_qualitative}

\begin{figure}[H]
    \centering
    \includegraphics[width=\linewidth]{figs/appendix_qualitative_1.png}
    \caption{Additional qualitative example.}
    \label{fig:appendix_qualitative_1}
\end{figure}

\begin{tcolorbox}[colback=gray!5,colframe=black,title=Human Reference Caption,
                  boxrule=0.5pt,arc=2pt,left=4pt,right=4pt,top=4pt,bottom=4pt]
A medium shot of a large, oval pumpkin with a carved face at night. The pumpkin is orange, with indented ridges running from the top to the bottom and a short, thick stem cut off at the top. The pumpkin has one triangle eye carved at the top center with a big open month [sic] smile carved below it. One small tooth on the top left side is part of the carved smile. The glow inside the pumpkin is golden. A small amount of dried leaves and dirt is visible inside at the bottom. An orange light coming from the left shines on the left side of the pumpkin and illuminates the green and dried plant life around it. A thick cord is visible in the background, horizontally behind the pumpkin.

\end{tcolorbox}

\vspace{0.4em}

\begin{tcolorbox}[colback=gray!5,colframe=black,title=QwenVL2.5-3B,
                  boxrule=0.5pt,arc=2pt,left=4pt,right=4pt,top=4pt,bottom=4pt]
The image shows a pumpkin that has been carved into a jack-o'-lantern. The pumpkin is illuminated from within, casting a warm glow. The carving features a simple face with a triangular nose and a small, round eye. The mouth is not visible in this view. The pumpkin is placed outdoors, surrounded by greenery and fallen leaves, indicating it might be in a garden or a park. The lighting suggests it is either dusk or nighttime, as there is a red hue to the overall image, possibly due to artificial lighting or the time of day. The pumpkin appears to be part of a Halloween decoration or event.
\end{tcolorbox}

\begin{tcolorbox}[colback=gray!5,colframe=black,title=CapRL-3B,
                  boxrule=0.5pt,arc=2pt,left=4pt,right=4pt,top=4pt,bottom=4pt]
The image depicts a creatively carved pumpkin, likely intended for Halloween decoration. Here is a detailed description: 1. **Object and Setting:** - The image shows a pumpkin that has been carved into a jack-o'-lantern, specifically designed to resemble a character with a yellow face. 2. **Pumpkin Description:** - The pumpkin is orange, typical of a standard Halloween pumpkin, and it appears to be a large, mature variety, possibly a pumpkin used for carving. 3. **Carving Details:** - The pumpkin has a triangular yellow triangle cut out for the top of the head, resembling a hat or hair, which is red and textured, likely made from paper or cardboard. This suggests the pumpkin might be styled as a character with hair, possibly a duck or a similar figure. 4. **Face Features:** - The pumpkin's face is yellow inside, illuminated by light, likely from an internal candle or LED light, given the red hue in the image. It features: - A single yellow triangle for the eyes, positioned at the top center. - An orange leaf on the right side, likely representing a nose or another facial feature, also cut from paper or cardboard. 5. **Body and Structure: - The pumpkin has a ribbed texture typical of pumpkins, with vertical ridges running down its body. 6. **Lighting:** - The image has a red tint, possibly due to a red light source, likely from a fire or a Halloween-themed lighting setup, common during the fall season. 7. **Location and Surroundings:** - The pumpkin is situated outdoors, surrounded by foliage with autumn leaves, indicating it is likely in a garden or park during fall. The ground is covered with fallen leaves, predominantly red, suggesting it is autumn. 8. **Leaves and Plants: - The pumpkin is surrounded by green plants and bushes with red leaves, typical of fall. There are some plants with thin stems and leaves, possibly small shrubs or vines. 9. **Additional Elements: - On the right side of the pumpkin, there is a small piece of paper or cardboard with the number "4" visible, possibly part of a shopping basket or a label, indicating this might be in a residential area or a community event. 10. **Position and Orientation: - The pumpkin is upright and facing forward, with its mouth open, showing the interior filled with some leaves, likely from the carving process. This detailed description should(truncated)
\end{tcolorbox}
\vspace{0.4em}

\begin{tcolorbox}[colback=gray!5,colframe=black,title=RubiCap-3B,
                  boxrule=0.7pt,arc=2pt,left=4pt,right=4pt,top=4pt,bottom=4pt]
The image depicts a creatively carved pumpkin, likely intended for Halloween decoration. The pumpkin has a traditional jack-o'-lantern face, with a triangular nose and a wide, open mouth that is filled with small, yellowish-orange leaves. The top of the pumpkin features a red, ribbed texture, resembling a hat or a hood, which adds an artistic touch to the carving. The pumpkin is placed outdoors, surrounded by fallen leaves and other foliage, suggesting it is in a garden or a yard during autumn. The lighting appears to be artificial, possibly from a nearby light source, casting a warm glow on the pumpkin and highlighting its features. The background is slightly blurred, focusing attention on the pumpkin itself. The overall scene conveys a festive and seasonal atmosphere, typical of Halloween celebrations.
\end{tcolorbox}

\begin{tcolorbox}[colback=gray!5,colframe=black,title=BalCapRL-3B,
                  boxrule=0.7pt,arc=2pt,left=4pt,right=4pt,top=4pt,bottom=4pt]
In the image, a pumpkin is prominently featured, illuminated by a warm, reddish light that casts a glow on its surface. The pumpkin has been carved to resemble a face, with a triangular nose made of a yellow material and a small, round eye cutout. The rest of the face is filled in with a dark orange or brown color, giving it a somewhat eerie appearance. The pumpkin is placed on a bed of fallen leaves, which are scattered around it, adding to the autumnal theme. The background is slightly blurred but shows more leaves and some greenery, suggesting that the pumpkin is outdoors, possibly in a garden or a park. The lighting creates a dramatic effect, highlighting the texture of the pumpkin's skin and the contrast between the carved features and the surrounding foliage.
\end{tcolorbox}

\begin{figure}[H]
    \centering
    \includegraphics[width=\linewidth]{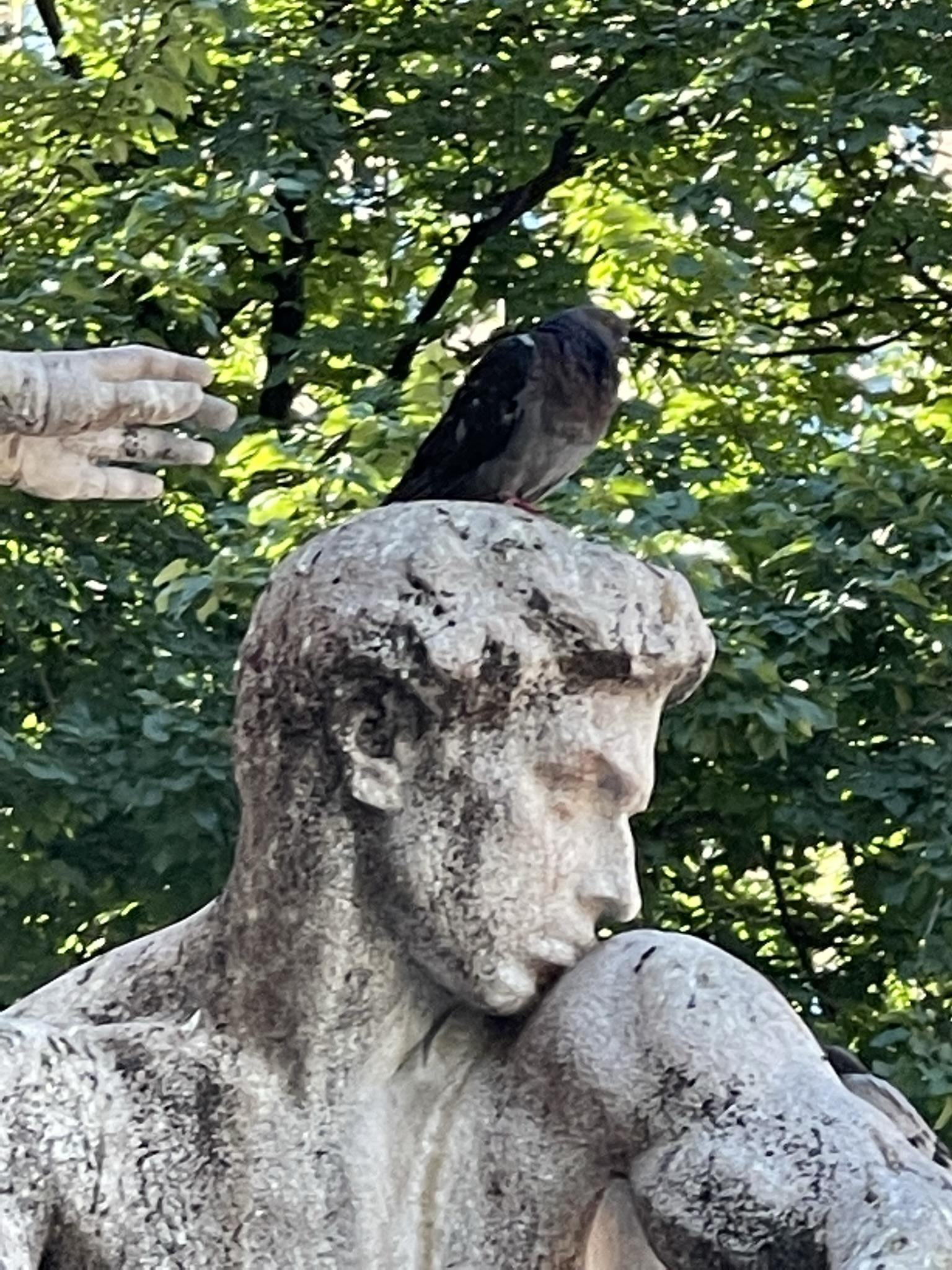}
    \caption{Additional qualitative example 2.}
    \label{fig:appendix_qualitative_2}
\end{figure}

\begin{tcolorbox}[colback=gray!5,colframe=black,title=Human reference caption,
                  boxrule=0.7pt,arc=2pt,left=4pt,right=4pt,top=4pt,bottom=4pt]
A front view of a stone statue that depicts a man from the chest up, turning his head to the right and his lip on his right shoulder. The man's nose is slightly above the shoulder, leaving a gap between the nose and the shoulder. The statue has patches of black dirt all over it. A black and gray bird is sat on top of the statue's head, facing toward the right of the image. A stone hand that is partially cut from the left of the image is reaching toward the bird. Multiple leaves from a tree are covering the background behind the statue. Sunlight is shining on the leaves.
\end{tcolorbox}

\begin{tcolorbox}[colback=gray!5,colframe=black,title=QwenVL2.5-3B,
                  boxrule=0.7pt,arc=2pt,left=4pt,right=4pt,top=4pt,bottom=4pt]
The image shows a bird perched on top of a statue's head. The statue appears to be of a human figure, possibly a woman, with detailed facial features and a serene expression. The bird is positioned near the top of the statue's head, looking upwards. The background consists of green foliage, suggesting that the statue is located in an outdoor setting, possibly a park or garden. The bird has a mix of brown and white feathers, and its wings are slightly spread out. The statue itself is weathered, indicating it has been there for some time.
\end{tcolorbox}

\begin{tcolorbox}[colback=gray!5,colframe=black,title=CapRL-3B,
                  boxrule=0.7pt,arc=2pt,left=4pt,right=4pt,top=4pt,bottom=4pt]
The image depicts a scene in what appears to be a park or garden, likely a historical or possibly a cemetery, given the presence of an old statue and a pigeon perched on it. Here is a detailed description: 1. **Setting and Subject:** - The image features an old, weathered statue of a human bust, likely from classical or ancient times, possibly Roman or Greek, given its style. The statue is carved from stone and shows signs of age and wear, with a rough, textured surface covered in dirt and moss, suggesting it is outdoors and possibly in a graveyard or a historical garden. 2. **Statue:** - The statue is of a person with a head that is prominently displayed in the foreground. The statue is of a man with a profile view, facing to the left. He has a rugged, aged face with a beard and mustache, characterized by a bald head with a brown spot, likely a birthmark or mark of age. The statue's hands are clasped together, holding something, though only part of one arm is visible, showing a stone arm with small black spots, possibly ants or insects. 3. **Pigeon: Perched on the Statue:** - A pigeon is perched on the top of the statue's head, specifically on the male figure's forehead. The pigeon has a brownish plumage with white markings on its chest and wings, typical of a rock dove or a common pigeon. 4. **Hand in the Background:** - On the left side of the image, there is a hand wearing gloves, likely belonging to a gardener or someone working in the area. The hand is raised, pointing towards the pigeon, possibly indicating that the person is either feeding the pigeon or gesturing at it. 5. **Environment:** - The background is lush with green foliage, likely trees or bushes, suggesting a park or forested area. The trees are dense and leafy, indicating it might be summer or spring due to the bright sunlight filtering through the leaves. 6. **Details of the Statue: - The statue's head is slightly tilted, and the face appears to be aged, with visible cracks and spots, typical of outdoor stone statues exposed to elements over time. 7. **Objects in the Image: - The statue is holding a stone object in its right arm, which could be a book or another item, but it is not clearly visible. 8. **Text and Numbers:** - There are no ..

\end{tcolorbox}

\begin{tcolorbox}[colback=gray!5,colframe=black,title=RubiCap-3B,
                  boxrule=0.7pt,arc=2pt,left=4pt,right=4pt,top=4pt,bottom=4pt]
The image depicts an outdoor scene featuring a weathered statue of a person. The statue appears to be made of stone or a similar material, showing signs of age and wear, including patches of discoloration and possible damage. The statue is positioned in a natural setting, surrounded by lush green foliage, indicating that it might be located in a park or garden. On top of the statue's head, there is a bird perched, possibly a pigeon or dove, given its small size and the typical appearance of such birds. The bird is facing away from the camera, looking towards the background. In the foreground, on the left side of the image, there is a human hand reaching out towards the statue. The hand is partially visible, with fingers extended as if trying to touch or interact with the statue. The skin tone of the hand suggests it belongs to a person who is not part of the statue itself. The overall atmosphere of the image is serene and somewhat contemplative, blending elements of nature and human artistry. The juxtaposition of the aged statue and the living bird creates a contrast between the static and dynamic aspects of the scene.
\end{tcolorbox}

\begin{tcolorbox}[colback=gray!5,colframe=black,title=BalCapRL-3B,
                  boxrule=0.7pt,arc=2pt,left=4pt,right=4pt,top=4pt,bottom=4pt]
In the image, a pigeon is perched on the top of a statue's head. The statue appears to be made of stone or concrete and features a human face with a somewhat weathered and aged appearance. The pigeon has a mix of brown and white feathers, with its wings slightly spread out. The statue's head is slightly tilted to the side, and the pigeon is positioned near the top of the head, close to the ear area. The background is lush with green foliage, indicating that the statue is likely located in an outdoor setting, possibly a park or garden. There is a hand visible in the foreground, wearing gloves, which seems to be reaching towards the pigeon. The hand is positioned to the left of the statue, and the fingers are extended as if trying to interact with the pigeon. The statue's shoulders and part of its upper body are also visible, showing more of the weathered texture of the stone.
\end{tcolorbox}

\subsection{Linear length penalty.}
\label{sec:linear_length}
\begin{equation}
\rho = \frac{\ell_{\mathrm{pred}}}{\ell_{\mathrm{ref}}},
\end{equation}
we define the deviation from the acceptable range $[\tau_l, \tau_u]$ as
\begin{equation}
d(\rho) = \max(\rho - \tau_u, 0) + \max(\tau_l - \rho, 0).
\end{equation}
The linear length penalty is then
\begin{equation}
p_{\mathrm{len}} = \lambda_{\mathrm{len}} \, d(\rho),
\end{equation}
where $\lambda_{\mathrm{len}}$ controls the penalty strength. This penalty is subtracted from the normalized advantage:
\begin{equation}
\tilde{A} = A - p_{\mathrm{len}},
\end{equation}
where $A$ is the normalized advantage and $\tilde{A}$ is the penalized advantage used for optimization.


\newpage

\end{document}